# یک رویکرد یادگیری انتقالی با شبکه عصبی کانولوشنال برای تشخیص افراد دارای ماسک از روی تصاویر

ابوالفضل یونسی ۱، دانشجو، رضا افروزیان ۲، استادیار، یوسف صیفاری۳، استادیار

۱- دانشکده فنی و مهندسی میانه - دانشگاه تبریز - میانه - ایران - younesi.abolfazl@yahoo.com
۲- دانشکده فنی مهندسی میانه - دانشگاه تبریز - میانه - ایران - afrouzian@tabrizu.ac.ir
۳- دانشکده فنی و مهندسی - دانشگاه مراغه - مراغه - ایران - seyfari@maragheh.ac.ir

**چکیده:** با توجه به همه‌گیری ویروس کرونا (کووید-۱۹) و انتقال سریع آن در سرتاسر دنیا، جهان با یک بحران بزرگ روبرو شده است. برای جلوگیری از شیوع ویروس کرونا سازمان بهداشت جهانی (WHO) استفاده از ماسک و رعایت فاصله اجتماعی در مکان‌های عمومی و شلوغ را بهترین روش پیشگیرانه معرفی کرده است. این مقاله یک سیستم برای شناسایی افراد دارای ماسک پیشنهاد می‌کند که بر پایه یادگیری انتقالی و معماری Inception v3 است. روش پیشنهادی با استفاده از دو مجموعه داده SMFD (Simulated Mask Face Dataset) و (MaskedFace-Net) MFN آموزش می‌بیند و با تنظیم بهینه فراپارامترها و طراحی دقیق بخش تماماً متصل سعی می‌کند دقت سیستم پیشنهادی را افزایش دهد. از مزایای سیستم پیشنهادی این است که می‌تواند علاوه بر صورت‌های دارای ماسک و بدون ماسک، حالت‌های استفاده غیر صحیح از ماسک را نیز تشخیص دهد. از این‌رو روش پیشنهادی تصاویر چهره ورودی را به سه دسته تقسیم‌بندی خواهد کرد. نتایج آزمایشی، دقت و کارایی بالای روش پیشنهادی را در موضوع فوق نشان می‌دهند؛ بطوری‌که این مدل در داده‌های آموزش به دقت ۹۹.۴۷٪ و در داده‌های آزمایشی به دقت ۹۹.۳۳٪ دست یافته است.

**واژه‌های کلیدی:** ماسک، کووید-۱۹، یادگیری انتقالی، شبکه عصبی کانولوشنال، معماری InceptionV3.

# A transfer learning approach with convolutional neural network for Face Mask Detection

Abolfazl Younesi, Student[1], Reza Afrouzian, Assitant Professor[2], Yousef Seyfari[3], Assistant Professor[3]

1- Miyaneh Faculty of Engineering, University of Tabriz, Miyaneh, Iran, Email: younesi.abolfazl@yahoo.com
2- Miyaneh Faculty of Engineering, University of Tabriz, Miyaneh, Iran, Email: afrouzian@tabrizu.ac.ir
3- Faculty of Engineering, University of Maragheh, Maragheh, Iran, Email: seyfari@maragheh.ac.ir

**Abstract:** Due to the epidemic of the coronavirus (Covid-19) and its rapid spread around the world, the world has faced a huge crisis. To prevent the spread of the coronavirus, the World Health Organization (WHO) has introduced the use of masks and keeping social distance as the best preventive method. So, developing an automatic monitoring system for detection of facemask in some crowded places is essential. To do this, we propose a mask recognition system based on transfer learning and Inception v3 architecture. In the proposed method, two datasets are used simultaneously for training including: Simulated Mask Face Dataset (SMFD) and MaskedFace-Net (MFN).this paper tries to increase the accuracy of the proposed system by optimally setting hyper-parameters and accurately designing the fully connected layers. The main advantage of the proposed method is that in addition to masked and unmasked face, it can also detect cases of incorrect use of mask. Therefore, the proposed method classifies the input face images into three categories. Experimental results show the high accuracy and efficiency of the proposed method; so that, this method has achieved to accuracy of 99.47% and 99.33% in training and test data respectively.

**Keywords:** Mask, Covid-19, Transfer Learning, convolutional neural network, Inception v3.

## ۱- مقدمه

قبل از شیوع ویروس کرونا (کووید-۱۹) عده کمی از مردم در جهت حفظ سلامت خود در برابر آلودگی هوا از ماسک استفاده می‌کردند. بعد از همه‌گیری کووید-۱۹ روند استفاده از ماسک برای جلوگیری از شیوع کووید-۱۹ در سراسر جهان افزایش یافت. مطالعات نشان می‌دهد که استفاده از ماسک تأثیر بسزایی در کاهش روند انتقال کووید-۱۹ دارد. به همین دلیل سازمان بهداشت جهانی (WHO) پیشنهاد می‌کند افرادی که در مکان‌های عمومی قرار می‌گیرند از ماسک صورت استفاده کنند [۱۹]. مطابق گزارشات WHO تا روز ۲ مهر ۱۴۰۰ حدود ۲۳۰ میلیون انسان مبتلا به ویروس کووید-۱۹ شناسایی شده‌اند و توصیه سازمان جهانی بهداشت برای جلوگیری برای افزایش تعداد مبتلایان به کرونا، واکسینه شدن، استفاده از ماسک و حفظ فاصله اجتماعی است. بااین‌وجود، استفاده کم افراد از ماسک (مخصوصاً در برخی از مکان‌های عمومی) منجر به کاهش رعایت پروتکل‌های بهداشتی می‌شود. به همین دلیل در برخی از مکان‌های عمومی بهتر است نظارت کافی بر استفاده از ماسک صورت پذیرد. در همین راستا، به دلیل سخت و هزینه‌بر بودن نظارت انسانی در استفاده صحیح از ماسک، می‌توان با استفاده از تکنیک‌های هوش مصنوعی، سیستمی را در حوزه بینایی ماشین طراحی کرد. سیستم فوق می‌تواند افراد دارای ماسک را از افراد بدون ماسک و یا حتی افرادی که به فرم صحیح از ماسک استفاده نکرده‌اند، تشخیص دهد.

در نتیجه تمامی موارد بالا باعث شده است که دولت‌ها شروع به اعمال جریمه برای افرادی که ماسک نزده‌اند نمایند تا از شدت انتقال ویروس بکاهند. رشد چشمگیر یادگیری عمیق در سال‌های اخیر باعث شده است که در پژوهش و صنعت در همه زمینه‌ها از یادگیری عمیق استفاده گردد. یکی از مهم‌ترین زمینه‌هایی که از یادگیری عمیق استفاده می‌شود، پردازش تصویر و تشخیص اشیا است [۲۷،۲۶]. با کمک یادگیری عمیق، موضوعاتی مثل تشخیص اشیا، کلاس‌بندی و تقسیم‌بندی تصاویر بادقت بسیار زیادی در حال انجام است. باتوجه‌به اینکه نظارت و شناسایی افراد دارای ماسک در حوزه بینایی ماشین و کلاس‌بندی است، می‌توان از یادگیری عمیق و مزایای آن در این حوزه نیز بهره برد. به همین منظور، در این مقاله از تکنیک یادگیری عمیق برای شناسایی و کلاس‌بندی افراد ماسک‌زده در فضاهای عمومی استفاده می‌شود. هدف اصلی این مقاله ارائه سیستمی مبتنی بر یادگیری عمیق برای شناسایی افراد دارای ماسک است تا بدین‌وسیله بتوان یک سیستم اتوماتیک بدین منظور طراحی کرد.

نوآوری‌های اصلی این مقاله به شرح ذیل است:

- بررسی طراحی‌های مختلف برای بخش دسته‌بندی شبکه و ارائه معماری بهینه برای آن همراه با شبکه Inception v3 که به‌منظور استخراج ویژگی مورد استفاده قرار می‌گیرد.
- به‌کارگیری هم‌زمان پایگاه‌داده‌های MaskedFace-Net(MFN) با Simulated Mask Face Dataset (SMFD) برای آموزش و ارزیابی شبکه پیشنهادی و دسته‌بندی تصاویر چهره به سه کلاس دارای ماسک، بدون ماسک و فرم ناصحیح.

روش پیشنهادی در این مقاله به‌گونه‌ای طراحی شده است که بتواند افراد دارای ماسک از افرادی که ماسک نزده‌اند و یا حتی ماسک را به گونه درست بر صورت خود قرار نداده‌اند، تشخیص دهد. این روش بر پایه یادگیری انتقالی و معماری Inception v3 توانسته این کار را در کمترین زمان ممکن و با بیشترین دقت انجام دهد.

بخش‌بندی ادامه مقاله به این صورت خواهد بود: در بخش دوم مقالات موجود در حوزه تشخیص افراد دارای ماسک مورد بررسی قرار می‌گیرد. بخش سوم ابتدا به توضیح دقیق در مورد پایگاه‌داده‌های مورداستفاده در مقاله می‌پردازد و سپس روش پیشنهادی با جزئیات مورد بررسی قرار می‌گیرد. در بخش چهارم نیز نتایج آزمایشی ارائه می‌شود و در نهایت، بخش آخر شامل نتیجه‌گیری و ارائه پیشنهاداتی برای کارهای آینده خواهد بود.

## ۲- مروری بر کارهای قبلی

یکی از کاربردهای مهم بینایی ماشین که در حال پیشرفت روزافزون است حوزه دسته‌بندی و تشخیص اشیا می‌باشد. به علت شیوع ویروس کرونا (کووید-۱۹) در سال‌های اخیر، محققان و تحلیلگران زیادی علاقه‌مند به استفاده از الگوریتم‌های بینایی ماشین در این حوزه شده‌اند و سعی شده است با استفاده از فناوری‌های تشخیص چهره و دسته‌بندی اشیا، الگوریتم‌هایی را ارائه دهند که به شناسایی افرادی دارای ماسک بپردازند [۴، ۶ و ۱۰]. به دلیل اینکه از شروع بیماری مربوط به همه‌گیری ویروس کرونا حدوداً یک سال و چندین ماه می‌گذرد، ازاین‌رو مقالات ارائه شده در این حوزه زیاد نیستند و فقط محدود به چند مقاله و کار تحقیقاتی است که در ماه‌های اخیر منتشر شده‌اند. در این بخش تعدادی از کارهای پیشین مرتبط به طور خلاصه مرور می‌شود.

جیانگ و همکاران در سال ۲۰۲۰ [۱۳] یک آشکارساز یک مرحله‌ای کارآمد به نام RetinaFaceMask را با استفاده از یک چارچوب مدل هرمی برای شناسایی ماسک صورت ارائه نمودند. این مدل برای اینکه عملکرد بهتری را ارائه دهد از مکانیزم توجه ویژگی حذف شی برای بیشتر شدن دقت استفاده می‌کند. دقت روش پیشنهادی ۹۳٪ می‌باشد و همچنین از نقاط قوت این مقاله می‌توان به پیاده‌سازی این مدل بر روی تلفن همراه اشاره کرد که از مدل از پیش‌آموزش‌دیده و کم‌حجم MobileNet استفاده می‌کند. در سال ۲۰۲۱ پریتی ناگارته و همکاران [۱۲] با استفاده از آشکارساز SSDMNV2 و MobileNetV2 یک سیستم تشخیص ماسک به نام SSDMNV2 با میانگین دقت ۹۳٪ ارائه نمودند. محمد لویی و همکاران در سال ۲۰۲۱ از ResNet-50 برای آموزش سیستم و از ماشین بردار پشتیبان جهت تشخیص چهره افرادی که از ماسک استفاده می‌کنند بهره بردند. آن‌ها در مجموعه‌داده چهره‌های واقعی دارای ماسک (RMFD) و مجموعه‌داده شبیه‌سازی

شده افرادی که ماسک زده‌اند (SMFD) به ترتیب به‌دقت ۹۹/۶۴٪ و ۹۹/۴۹٪ دست یافتند [۱۱].

در همین سال فوزیه ایم ای‌اوکور و همکارانش [۱۵] از مجموعه‌داده آزمایشگاه‌های سیستم تشخیص ماسک صورت بدون محدودیت ISL-UMFD استفاده کردند و با سه معماری مدل ResNet-50، Inception v3 و MobileNetV2 به ترتیب به دقت‌های ۹۵/۶۳٪، ۹۸/۲۰٪ و ۹۷/۹۱٪ دست یافتند. در سال ۲۰۲۱ محمد لویی و همکارانش [۴] در مقاله دیگری مجموعه‌داده Kaggle-medical mask را که از ترکیب دو مجموعه‌داده عمومی Medical Masks Dataset (MMD) و Face Mask Dataset (FMD) است، استفاده کردند که با معماری ResNet-50 به‌عنوان مدل از پیش آموزش‌دیده و از YOLOv2 برای تشخیص ماسک استفاده نمودند و به‌دقت متوسط ۸۱٪ دست یافته‌اند. میلیتنت و همکاران در سال ۲۰۲۰ با استفاده از معماری VGG16 توانستند به‌دقت ۹۶٪ در تشخیص ماسک دست یابند [۲]. این مقاله بر روی بردهای رزبری‌پای پیاده‌سازی شده است.

صابر اجاز و همکارش نیز در سال ۲۰۱۹ با استفاده از معماری از پیش آموزش‌دیده Face-net، برای استخراج ویژگی، هشت حالت متفاوت را بر روی سه مجموعه‌داده مختلف AR، Mask Face Dataset و IIIT v1 بررسی نموده و به ترتیب در بهترین حالت و حالت متوسط به‌دقت ۹۸/۱٪ و ۸۲/۴٪ دست یافتند [۱۶]. آنها از ماشین بردار پشتیبان[1] (SVM) برای طبقه‌بندی تصاویر استفاده کردند. رویکردی که آنها داشتند یک رویکرد دو مرحله‌ای بوده‌است. ابتدا تشخیص ماسک را با MTCNN انجام داده‌اند و سپس تشخیص چهره را در مرحله دوم انجام داده‌اند.

باتوجه به موارد فوق به این نتیجه می‌رسیم که معماری‌های مختلف و نحوه تنظیم فراپارمترهای[2] آن‌ها بر روی یک مجموعه‌داده می‌تواند تفاوت‌های قابل‌توجهی را در دقت سیستم‌ها ایجاد کند. معماری Inception v3 توانسته است در مصالحه[3] بین تعداد پارامترها و دقت تشخیص به یک توازن نسبتاً خوبی دست پیدا کند. ازاین‌رو، روش پیشنهادی در این مقاله نیز از معماری Inception v3 برای استخراج ویژگی استفاده می‌کند. از طرفی، اکثر روش‌های مربوط به مقالات قبلی تنها خروجی دو کلاس (چهره دارای ماسک و بدون ماسک) داشته‌اند که در کاربردهای عملی، علاوه بر دو کلاس فوق، نیاز است که سیستم همچنین توانایی تشخیص چهره‌هایی را داشته باشد که ماسک را به فرم ناصحیح استفاده می‌کنند. ازاین‌رو در این مقاله سعی شده است که به این موضوع توجه ویژه‌ای گردد و لایه‌های تماماً متصل قرار گرفته در خروجه شبکه Inception v3 به‌نوعی طراحی گردند تا روش پیشنهادی توان دسته‌بندی تصاویر ورودی به سه کلاس فوق، را داشته باشد. همچنین این مقاله بر اساس پایگاه‌داده‌های در دسترس، به‌منظور آموزش درست و دقیق، از ترکیب دو مجموعه‌داده MFN و SMFD استفاده می‌کند.

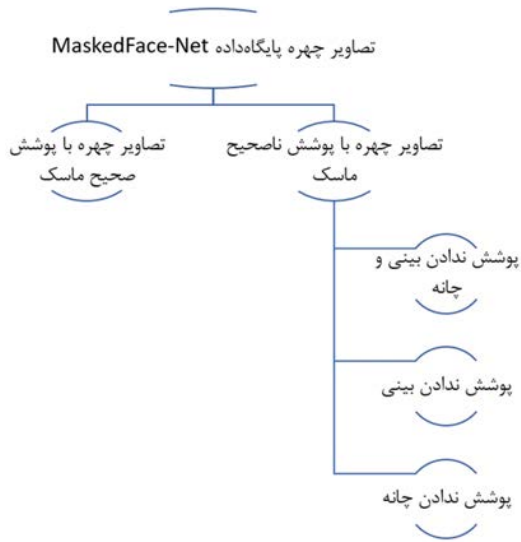

**شکل ۱) ساختار مجموعه‌داده Masked Face-Net**

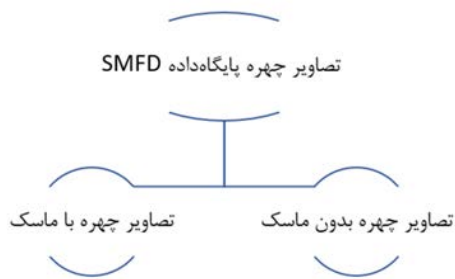

**شکل ۲) ساختار مجموعه‌داده SMFD**

## ۳- روش پیشنهادی

در این مقاله برای دسته‌بندی چهره افراد، یک مدل بر مبنای یادگیری انتقالی عمیق پیشنهاد شده است. در ادامه به بررسی جزئیات معماری مدل پیشنهادی و دیگر بخش‌های مدل خواهیم پرداخت.

### ۳-۱- مجموعه‌داده

مجموعه‌داده‌های استفاده شده در این مقاله شامل دو مجموعه‌داده MFN و SMFD است [۱،۳]. مجموعه‌داده MFN یک مجموعه‌داده مصنوعی است که دارای ۱۳۷۰۱۶ تصویر می‌باشد. تصاویر این مجموعه‌داده به دودسته تقسیم می‌گردد: ۱) افرادی که به‌صورت صحیح ماسک زده‌اند (CMFD) و ۲) افرادی که به طور نادرست از ماسک استفاده کرده‌اند (IMFD).

دسته دوم خود به سه زیر دسته تقسیم می‌شوند که عبارت‌اند از: ۱) چانه بدون پوشش، ۲) بینی بدون پوشش و ۳) بینی و دهان بدون پوشش. باتوجه به درنظرگیری حالات مختلف استفاده غیرصحیح از ماسک، مجموعه‌داده شامل جزئیات بیشتری برای آموزش و تشخیص

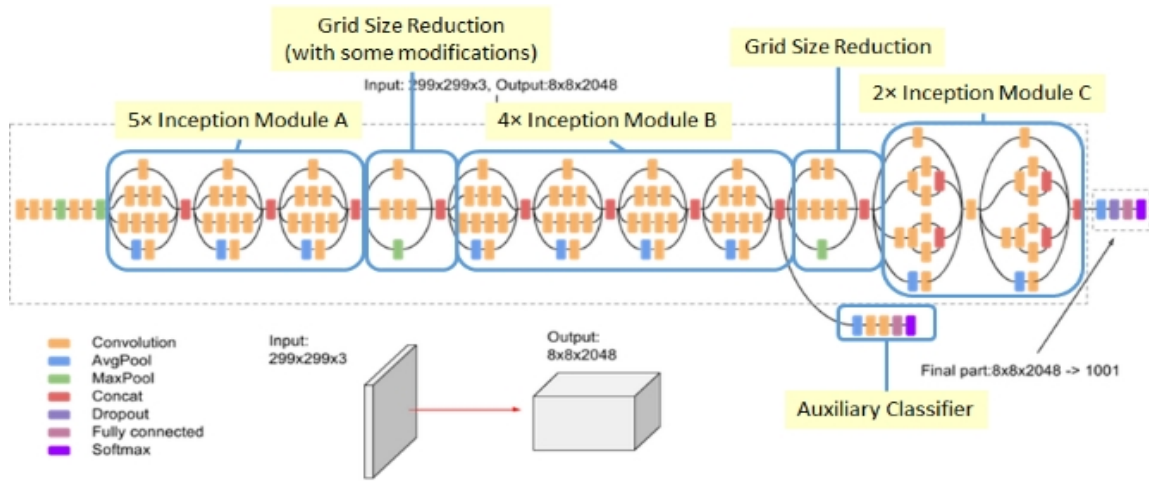

**شکل ۳) معماری مدل از پیش آموزش دیده Inception v3 [۱۸،۲۲]**

است. در شکل ۱ ساختار دسته‌های مجموعه‌داده MFN مشاهده می‌شود. مجموعه‌داده SMFD نیز دارای ۱۵۷۰ تصویر در دو بخش افرادی که ماسک دارند و افرادی که ماسک ندارند می‌باشد که ساختار آن در شکل ۲ مشاهده می‌شود.

به دلیل اینکه هدف مقاله دسته‌بندی تصاویر چهره ورودی به سه کلاس دارای ماسک، بدون ماسک و فرم ناصحیح است، ازاین‌رو از تصاویر هر دو پایگاه‌داده به طور هم‌زمان برای آموزش، اعتبارسنجی و تست استفاده شده است.

برای دسته‌بندی تصاویر ورودی، ابتدا تصویر وارد یک سیستم آشکارسازی چهره می‌شود تا بخش ناحیه صورت را در تصویر شناسایی کند. بدین منظور از روش viola jones جهت آشکارسازی ناحیه صورت در تصویر استفاده شده است [۲۰]. سپس ناحیه چهره به‌عنوان ورودی به شبکه پیشنهادی داده می‌شود تا بعد از انجام پردازش‌های لازم آن را به سه دسته دارای ماسک، بدون ماسک و استفاده ناصحیح از ماسک طبقه‌بندی کند که به ترتیب با کادر مستطیلی رنگ سبز، قرمز و نارنجی در ناحیه صورت از تصویر مشخص می‌شوند.

### ۳-۲- معماری شبکه

همان‌طور که قبلاً نیز بحث شد، برای استخراج ویژگی در شبکه پیشنهادی از معماری Inception v3 استفاده شده است. همچنین روش پیشنهادی سعی در یک طراحی مناسب در بخش دسته‌بندی با استفاده از لایه‌های تماماً متصل[4] است تا عمل دسته‌بندی به سه کلاس مذکور را بادقت بالایی انجام دهد. در شکل ۳ معماری Inception و در شکل ۴ بلوک دیاگرام ساختار پیشنهادی ارائه شده است.

مدل Inception یک شبکه عصبی کانولوشن است که معمولاً برای کاربردهای از نوع دسته‌بندی اشیاء مورد استفاده قرار می‌گیرد. همچنین به عنوان GoogLeNet نیز شناخته می‌شود. شبکه مذکور از مجموعه داده‌های ImageNet برای فرآیند آموزش استفاده می‌کند و

اولین بار در سال ۲۰۲۱ معرفی شد. لایه های آن ترکیبی از لایه کانولوشن ۱×۱، ۳×۳، ۵×۵ و ۷×۷ است که بانک های فیلتر خروجی آنها در یک واحد به هم متصل شده‌اند. لایه کانالوشنی ۷×۷ به شبکه Inception V3 اضافه شده است و می‌تواند ویژگی‌ها را در چند سطح استخراج کند. به عبارتی یک استخراج کننده ویژگی چندسطحی است [۲۱،۲۲]. مدل از پیش آموزش دیده Inception V3 تعداد پارامترهای کمتر، دقت بالاتر (در پایگاه‌داده اعتبار سنجی ImageNet) و حجم کمتری در مقایسه با مدل‌های VGG16، VGG19 و Resnet50V2 دارد و همچنین در مقایسه با مدل‌های MobileNetV2 و DenseNet (در پایگاه‌داده اعتبار سنجی ImageNet) دقت بالاتری دارد [۲۴].

برای پیاده‌سازی روش فوق از کتابخانه Keras و TensorFlow استفاده شده است. یکی از چالش‌های این مسئله وجود تصاویر با اندازه‌های مختلف است که برای حل آن اندازه تمامی تصاویر ورودی به ۲۹۹×۲۹۹ تغییر می‌یابد که سایز پیش‌فرض معماری Inception v3 است [۲۵]. حجم تصاویر استفاده شده از کل مجموعه‌داده‌ها به‌منظور آموزش، اعتبارسنجی[5] و تست به ترتیب ۷۰٪، ۱۵٪ و ۱۵٪ در نظر گرفته شده است. همچنین در حین آموزش شبکه، تکنیک افزایش داده برای جلوگیری از بیش برازش مورد استفاده قرار می‌گیرد.

در روش پیشنهادی، همان‌طور که توضیح داده شد، از شبکه پیش آموزش‌دیده Inception v3 برای استخراج ویژگی استفاده می‌شود. سپس خروجی شبکه فوق به‌عنوان ورودی به لایه ادغام متوسط[6] داده می‌شود که اساساً هدف، کاهش اندازه داده‌ها است و سپس خروجی با استفاده از لایه مسطح[7] به یک بردار یک بعدی تبدیل می‌گردد. بردار فوق وارد لایه‌های مخفی می‌شود که بخشی از شبکه عصبی تماماً متصل جهت دسته‌بندی داده‌های ورودی است. لایه حذف تصادفی[8] نیز به‌منظور جلوگیری از بیش‌برازش مورد استفاده قرار می‌گیرد. در آخر، خروجی لایه فوق به لایه بیشینه هموار[9] جهت دسته‌بندی تحویل داده می‌شود.

وزن‌های از پیش آموزش‌دیده معماری Inception V3 بر اساس پایگاه‌داده ImageNet می‌باشد. برای آموزش شبکه از تابع بهینه‌ساز ADAM استفاده شده و از ۶۰ دوره تناوب[10] برای برازش[11] استفاده شده است. تابع خطا استفاده شده در روش فوق آنتروپی متقاطع طبقه‌ای[12] بوده و از معیار دقت برای ارزیابی سیستم استفاده شده است.

### ۳-۳- افزایش تصویر

افزایش داده[13] یکی از تکنیک‌هایی است که می‌توان برای جلوگیری از بیش برازش[14] استفاده کرد. به‌طورکلی می‌توان گفت که روش افزایش داده، یک سری داده جدید به‌صورت مصنوعی بر پایه مجموعه‌داده اصلی ایجاد می‌کند که باعث بهبود دقت و کاهش خطا می‌گردد [۷]. همچنین برای ایجاد یک مدل یادگیری عمیق دقیق، مدل باید به‌خوبی آموزش ببیند و به همین منظور باید تا حد ممکن دارای خطای اعتبارسنجی پایینی باشد. تقویت مجموعه‌داده به ما کمک می‌کند تا مجموعه‌داده جامع‌تری داشته باشیم [۵].

برای دستیابی به هدف فوق، با ایجاد یکسری تغییرات روی تصاویر مجموعه‌داده، از جمله انجام چرخش[15]، بزرگ‌نمایی[16]، تغییر در رنگ و شیفت[17] به بالا یا پایین، تصاویر جدیدی تولید شده و اندازه مجموعه‌داده افزایش یافته است [۶].

### ۳-۴- یادگیری انتقالی

یادگیری انتقالی[18] به این معنی است که می‌توان از ویژگی‌ها و وزن‌های مدل آموزش دیده برای آموزش دادن یک مدل جدید استفاده کرد [۱۰،۱۷]. با یادگیری انتقالی، اساساً سعی می‌کنیم از آموخته‌های خود در یک شبکه آموزش دیده برای یادگیری شبکه جدید استفاده کنیم. در این حالت اطلاعات وزن‌های شبکه قبلی به عنوان پارامترهای اولیه به شبکه فعلی انتقال داده می‌شوند. به عبارت دیگر، این روش اجازه می‌دهد که از توزیع‌های متفاوت برای آموزش و آزمایش استفاده شود و از آنجایی که شبکه عصبی کانولوشنی ویژگی‌های مهم یک تصویر را استخراج می‌کند، استفاده از یک مجموعه داده مناسب برای آموزش

اولیه، بخش مهمی از یک آموزش موفق را تشکیل می‌دهد. از این‌رو، در این روش فرایند یادگیری از ابتدا و با وزن‌های تصادفی شروع نمی‌شود بلکه با آغاز از الگوهای یادگرفته از حل یک مسئله متفاوت آغاز می‌شود و امکان ایجاد یک مدل صحیح را فراهم می‌آورد [۸،۹].

برای این‌کار، ابتدا وزن‌ها و پارامترهای مربوط به شبکه آموزش دیده پایه Inception V3 منجمد[19] می‌شوند و فرایند آموزش و بروزرسانی وزن‌ها تنها برای لایه‌های تماماً متصل بخش دسته‌بند اعمال می‌شود. بعد از آموزش لایه‌های تماماً متصل، وزن‌ها و پارامترهای مربوط به لایه‌های آخر شبکه پایه Inception v3 از حالت انجماد خارج شده و فرایند آموزش دوباره ادامه می‌یابد. بدین طریق، وزن‌های لایه‌های آخر شبکه پایه بر مبنای مسئله مورد نظر که در این مقاله تشخیص ماسک است، به‌طور دقیق تنظیم[20] می‌شود. از این‌رو، چالش‌هایی همچون

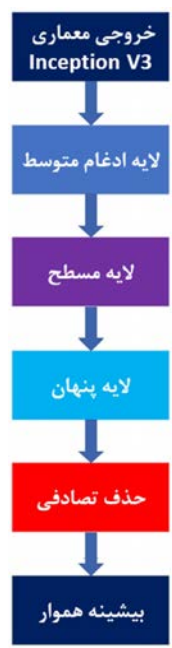

**شکل ۴) معماری شبکه مدل پیشنهادی**

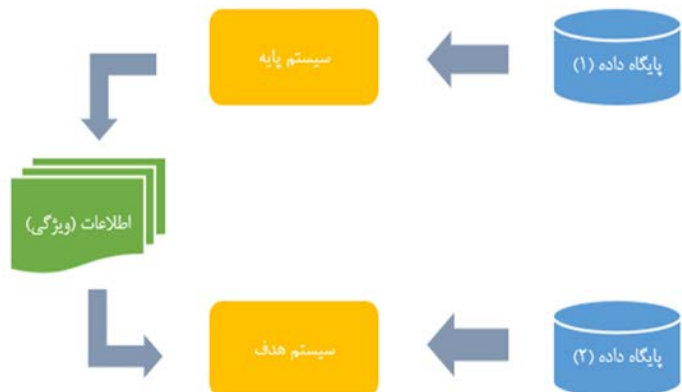

**شکل ۵) ساختار یادگیری انتقالی با استفاده از اطلاعات مربوط به سیستم پایه و به‌کارگیری آن‌ها در سیستم هدف، دیگر نیازی به آموزش سیستم هدف از ابتدا نخواهد بود.**

تعداد داده‌های محدود پایگاه داده و زمان مورد نیاز برای فاز آموزش را می‌توان با استفاده از تکنیک یادگیری انتقالی مدیریت کرد [۲۳].

در شکل ۵ ساختار یادگیری انتقالی نمایش داده شده است. همان‌طور که مشاهده می‌شود آموزش‌های انجام یافته بر روی سیستم پایه (از طریق پایگاه داده اول) به سیستم هدف منتقل شده و سیستم هدف فرایند آموزش را از ابتدا شروع نمی‌کند؛ بلکه با استفاده از اطلاعات پایگاه داده ۲، اطلاعات دریافتی را برای هدف خاصی بروزرسانی می‌نماید.

## ۴- نتایج روش پیشنهادی

برای ارزیابی مدل پیشنهادی از یک سیستم مجهز به واحد پردازنده مرکزی (CPU) Core i7-8850U با رم ۱۶ گیگابایت و کارت گرافیکی Nvidia GeForce MX150 با حافظه گرافیکی چهار گیگابایت استفاده شده است.

معیارهای مختلفی برای ارزیابی عملکرد روش پیشنهادی وجود دارد؛ مانند معیار دقت که مقدار آن برابر با تعداد مواردی است که سیستم به‌درستی پیش‌بینی کرده است و معمولاً این معیار برای مقایسه بین مدل‌ها مورداستفاده قرار می‌گیرد. در این مقاله علاوه بر معیار دقت از معیارهای یادآوری[۲۱]، دقت و میانگین هارمونیک یادآوری و دقت (f1-score) استفاده شده است. معیارهای ارزیابی استفاده شده توسط روابط زیر بیان می‌گردد:

$$Accuracy = \frac{TP+TN}{(TP+FP)+(TN+FN)} \quad (1)$$

$$\text{Re}call = \frac{TP}{(TP+FN)} \quad (2)$$

$$\Pr ecision = \frac{TP}{(TP+FP)} \quad (3)$$

$$F1Score = 2*\frac{\Pr ecision*\text{Re}call}{(\Pr ecision+\text{Re}call)} \quad (4)$$

در این مقاله برای رسیدن به پاسخ‌های مطلوب، طراحی‌های مختلفی برای بخش دسته‌بند که شامل لایه‌های تماماً متصل هستند، صورت‌گرفته است که نتایج روش پیشنهادی در ۷ حالت زیر مورد بررسی قرار می‌گیرد:

- ۳۲ نرون در یک‌لایه مخفی
- ۳۲ نرون در هر لایه مخفی (دارای دولایه مخفی)
- ۶۴ نرون در یک‌لایه مخفی
- ۶۴ نرون در هر لایه مخفی (دارای دولایه مخفی)
- ۱۲۸ نرون در یک‌لایه مخفی
- ۱۲۸ نرون در هر لایه مخفی (دارای دولایه مخفی)
- ۱۲۸ نرون در هر لایه مخفی (دارای سه‌لایه مخفی)

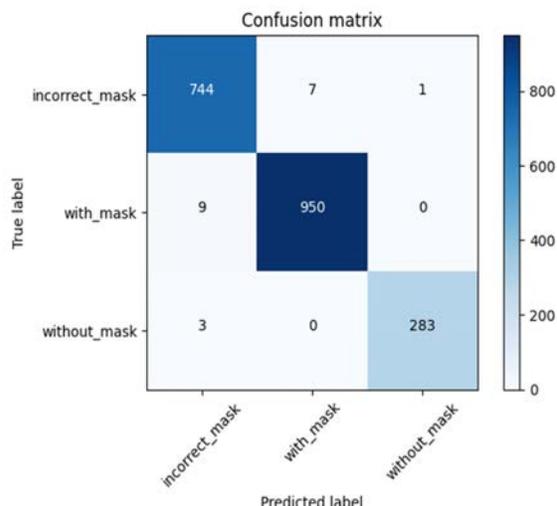

**شکل ۶) ماتریس سردرگمی با ۱۲۸ نرون در هر دو لایه مخفی**

**جدول ۱) گزارش طبقه‌بندی با ۱۲۸ نرون در هر دو لایه مخفی**

|  | Precision | Recall | F1-score |
| --- | --- | --- | --- |
| Incorrect_mask | 98.9 | 98.5 | 98.7 |
| With_mask | 99.3 | 99.3 | 99.1 |
| Without_mask | 99.8 | 99.5 | 99.1 |
|  |  |  |  |
| Accuracy |  |  | 99.4 |
| Macro_avg | 98.9 | 98.7 | 98.9 |
| Weighted_avg | 98.7 | 98.8 | 98.7 |

جدول ۲ مقایسه بین نتایج به‌دست‌آمده برای هفت حالت مختلف را مورد بررسی قرار می‌دهد. همان‌طور که مشاهده می‌شود بهترین نتایج در حالتی حاصل شده است که بخش دسته‌بند دارای دو لایه مخفی با ۱۲۸ نرون در هر لایه باشد (نتایج دقیق‌تر برای حالت مذکور در جدول ۱ نشان‌داده‌شده است). شکل ۷ نمودار دقت و میزان خطا را برای حالت فوق نشان می‌دهد. تعداد دوره تناوب برای آموزش شبکه، در حالتی که وزن‌های معماری Inception v3 منجمد هستند، برابر با ۴۰ در نظر گرفته شده و سپس به مدت ۲۰ بار در حالت غیرمنجمد برای لایه‌های آخر معماری Inception v3 تکرار می‌شود. در شکل ۶ نیز ماتریس سردرگمی[۲۲] به‌صورت گرافیکی نشان‌داده‌شده است.

ماتریس فوق اطلاعات دقیق‌تری را در مورد عملکرد سیستم پیشنهادی و خطاهای رخ‌داده نشان می‌دهد. همان‌طور که در شکل ۶ نشان‌داده‌شده است، بیشترین خطای دسته‌بندی مابین کلاس دارای ماسک و کلاس فرم ناصحیح است. دقت و توانایی سیستم پیشنهادی برای تشخیص افراد دارای ماسک ۹۹/۳ درصد، بدون ماسک ۹۹/۸ درصد و فرم ناصحیح ۹۸/۹ درصد می‌باشد.

در جدول ۳ نیز مقایسه روش پیشنهادی با دیگر کارهای ارائه شده در این حوزه نمایش داده می‌شود. البته به دلیل اینکه کارهای مختلف از یک پایگاه‌داده یکسان برای آموزش و ارزیابی شبکه استفاده نمی‌کنند، شاید انجام مقایسه، کار منصفانه‌ای نباشد. بااین‌وجود، برخی از کارها با اینکه (به‌جز روش معرفی شده در [۱۵]) تصاویر ورودی را به‌جای سه دسته، به دو دسته کاملاً مجزا شامل ماسک و بدون ماسک

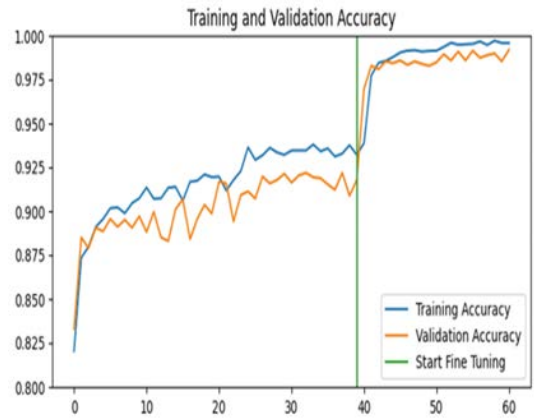

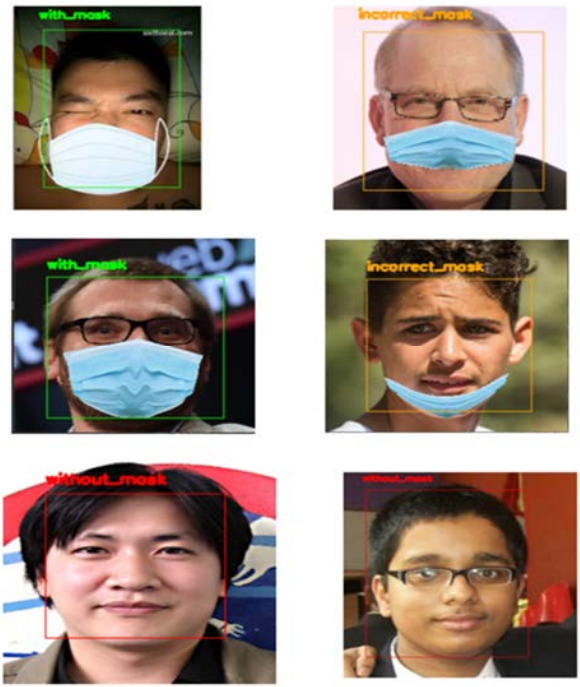

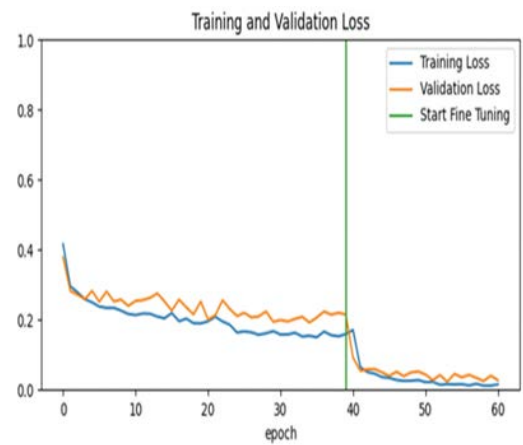

**شکل ۸) نمونه‌هایی از خروجی مدل پیشنهادی برای سه دسته مختلف**

**شکل ۷) الف) میزان تغییرات دقت با ۱۲۸ نرون در هر دو لایه مخفی**
**ب) میزان تغییرات خطا با ۱۲۸ نرون در هر دو لایه مخفی**

طبقه‌بندی می‌کنند (که دارای چالش کمتری است)، دقتی نسبتاً برابر با روش پیشنهادی دارند. با بررسی تصاویر پایگاه داده می‌توان دریافت که شباهت زیادی مابین کلاس استفاده غیرصحیح و دسته ماسک وجود دارد و زمانی که دسته‌بندی به سه کلاس انجام می‌شود، دقت سیستم علی‌رغم افزایش قابلیت آن، احتمالاً کاهش پیدا خواهد کرد.

الگوریتم ارائه شده در [۳] برای تشخیص افراد دارای ماسک، همچون روش پیشنهادی این مقاله، از یادگیری انتقالی بر مبنای معماری پایه Inception v3 استفاده می‌کند. بااین‌حال هر دو روش دارای تفاوت‌هایی هستند. اول اینکه روش پیشنهادی، همان‌طور که توضیح داده شد، چهره افراد را به سه دسته دسته‌بندی می‌کند درصورتی‌که خروجی مقاله مذکور تنها دارای دو دسته ماسک و بدون ماسک است.

روش پیشنهادی برای رسیدن به هدف فوق، در فرایند آموزش، علاوه بر پایگاه‌داده SMFD (پایگاه‌داده به‌کاررفته در [۳])، از پایگاه‌داده دیگری بنام MFN نیز استفاده می‌نماید (پایگاه‌داده فوق همان‌طور که قبلاً توضیح داده شد تصاویر چهره‌هایی که به فرم ناصحیح از ماسک استفاده می‌نمایند را نیز شامل می‌شود).

**جدول ۲) عملکرد روش پیشنهادی در حالت‌های مختلفی که برای بخش دسته‌بند شبکه در نظر گرفته شده است.**

| Neuron# | Train accuracy | Val accuracy | Train loss | Val loss | Image size | Layer | Epoch |
|---|---|---|---|---|---|---|---|
| ۳۲ | 98.8% | 98.6% | 2.2% | 2.4% | 299×299 | 1 | 60 |
| ۳۲ | 99.0% | 98.9% | 2.2% | 1.4% | 299×299 | 2 | 60 |
| ۶۴ | 99.1% | 98.9% | 2.3% | 3.8% | 299×299 | 1 | 60 |
| ۶۴ | 99.3% | 99.1% | 1.2% | 2.4% | 299×299 | 2 | 60 |
| ۱۲۸ | 99.3% | 98.8% | 1.2% | 3.6% | 299×299 | 1 | 60 |
| ۱۲۸ | 99.47% | 99.2% | 1.4% | 2.5% | 299×299 | 2 | 60 |
| ۱۲۸ | 99.5% | 98.7% | 1.3% | 2.3% | 299×299 | 3 | 60 |

جدول ۳) مقایسه کارهای مختلف ارائه شده برای تشخیص ماسک با روش ارائه شده در این مقاله (نتایج عددی مربوط به دقت ارائه شده برگرفته از مقالات مربوطه است)

| | Accuracy | Architecture | Dataset | Year |
|---|---|---|---|---|
| [12] | 93% | MobilenetV2 | Kaggle-medical mask | 2021 |
| [11] | 99.63% | ResNet-50 | RMFD | 2021 |
| [11] | 99.49% | ResNet-50 | SMFD | 2021 |
| [15] | 95.63% | ResNet-50 | ISL-UMFD | 2021 |
| [15] | 98.20% | MobilenetV2 | ISL-UMFD | 2021 |
| [15] | 97.91% | InceptionV3 | ISL-UMFD | 2021 |
| [4] | 81% | ResNet-50 | Kaggle-medical mask | 2021 |
| [2] | 96% | VGG16 | ----- | 2020 |
| [16] | 98% | Face net | IIIT version 1 | 2019 |
| [3] | 99.9% | Inception v3 | SMFD | 2020 |
| Proposed model | 99.33% | Inception v3 | MFN, SMFD | 2021 |

از طرف دیگر، الگوریتم پیشنهادی در این مقاله معماری‌های مختلف بخش دسته‌بند (شامل تعداد لایه‌های مختلف با تعداد نرون‌های متفاوت) را بررسی می‌کند تا بهترین طراحی ممکن را برای هدف موردنظر انجام دهد که با دسته‌بند مقاله مذکور دارای تفاوت‌هایی است. شکل ۸ چند نمونه از خروجی الگوریتم پیشنهادی را برای تصاویر ورودی از هر دسته نشان می‌دهد.

## ۵- نتیجه‌گیری و کارهای آینده

در این مقاله یک روشی با کمک یادگیری انتقالی و شبکه عصبی کانولوشنال Inception v3، برای دسته‌بندی افراد به سه کلاس ماسک، بدون ماسک و فرم ناصحیح پیشنهاد شده است. شبکه فوق از مجموعه‌داده MFN و SMFD برای آموزش و ارزیابی استفاده می‌کند. مدل پیشنهادی این قابلیت را دارد که هم بر روی تصویر و هم بر روی ویدئو عمل تشخیص را انجام دهد. برای افزایش دقت روش پیشنهادی، طراحی‌های مختلفی برای بخش دسته‌بند مدل با تعداد لایه‌ها و نرون‌های مختلف بررسی شده است تا حالت بهینه مشخص گردد. در نهایت روش پیشنهادی توانسته در بهترین حالت در مرحله آموزش به‌دقت ۹۹.۴۷ درصد و در مرحله آزمایش نیز به‌دقت ۹۹.۳۳ درصد با استفاده از معماری Inception v3 و تکنیک افزایش داده دست یابد.
از مواردی که در آینده می‌شود مورد بررسی قرار داد، می‌توان به طراحی سیستم‌هایی اشاره کرد که در جهت هوشمندسازی رعایت پروتکل‌های بهداشتی، تشخیص ماسک صورت و رعایت فاصله‌گذاری اجتماعی را به‌صورت هم‌زمان انجام می‌دهند.

## مراجع

[1] A. Cabani, K. Hammoudi, H. Benhabiles, and M. Melkemi, "MaskedFace-Net–A dataset of correctly/incorrectly masked face images in the context of COVID-19." Smart Health, vol.19, 2021.

[2] S. V. Militante and N. V. Dionisio, "Real-Time Facemask Recognition with Alarm System using Deep Learning," 2020 11th IEEE Control and System Graduate Research Colloquium (ICSGRC), pp. 106-110. IEEE, 2020

[3] G. J. Chowdary, N. S. Punn, S. K. Sonbhadra, and S. Agarwal, "Face mask detection using transfer learning of InceptionV3," International Conference on Big Data Analytics, pp. 81-90 Springer, Cham, 2020.

[4] M. Loey, G. Manogaran, M. Hamed, N. Taha, N. Eldeen, and M. Khalifa, Fighting Against COVID-19: A Novel Deep Learning Model Based on YOLOv2 with ResNet-50 for Medical Face Mask Detection" Sustainable cities and society, vol.65, 2021

[5] C. Shorten and T. M. Khoshgoftaar, "A survey on image data augmentation for deep learning," J. Big Data, vol. 6, no. 1, 2019.

[6] A. Mikolajczyk and M. Grochowski, "Data augmentation for improving deep learning in image classification problem," In 2018 international interdisciplinary PhD workshop (IIPhDW), pp. 117-122. IEEE, 2018.

[7] S. C. Wong, A. Gatt, V. Stamatescu, and M. D. McDonnell, "Understanding data augmentation for classification: When to warp?," In 2016 international conference on digital image computing: techniques and applications (DICTA), pp. 1-6. IEEE, 2016.

[8] W. M. Kouw and M. Loog, "An introduction to domain adaptation and transfer learning," arXiv preprint arXiv:1812.11806, 2018.

[9] S. J. Pan and Q. Yang, "A survey on transfer learning," IEEE Trans. Knowl. Data Eng., vol. 22, no. 10, pp. 1345–1359, 2010.

[10] M. Hussain, J. J. Bird, and D. R. Faria, "A study on CNN transfer learning for image classification In UK Workshop on computational Intelligence, pp. 191-202. Springer, Cham, 2018.

[11] M. Loey, G. Manogaran, M. H. N. Taha, and N. E. M. Khalifa, "A hybrid deep transfer learning model with machine learning methods for face mask detection in the era of the COVID-19 pandemic," Measurement (Lond.), vol. 167, no. 108288, p. 108288, 2021.

[12] P. Nagrath, R. Jain, A. Madan, R. Arora, P. Kataria, and J. Hemanth, "SSDMNV2: A real time DNN-based face mask detection system using single shot multibox detector and MobileNetV2," Sustain. Cities Soc., vol. 66, no. 102692, p. 102692, 2021.

[13] M. Jiang, F. Xinqi, and Y. Hong. "Retinamask: A face mask detector." arXiv preprint arXiv:2005.03950, 2020.

[14] N. Ud Din, K. Javed, S. Bae, and J. Yi, "A novel GAN-based network for unmasking of masked face," IEEE Access, vol. 8, pp. 44276–44287, 2020.

[15] F. I. Eyiokur, H. K. Ekenel, and A. Waibel, "Unconstrained face-mask & face-hand datasets: Building a computer vision system to help prevent the transmission of COVID-19," arXiv preprint arXiv:2103.08773, 2021.

[16] M. S. Ejaz and M. R. Islam, "Masked Face Recognition Using Convolutional Neural Network," 2019 International Conference on Sustainable Technologies for Industry 4.0 (STI), pp. 1-6, 2019

[17] J. Yosinski, J. Clune, Y. Bengio, and H. Lipson, "How transferable are features in deep neural networks?," Advances in neural information processing systems, vol. 27, 2014.

[18] D. Vasan, M. Alazab, S. Wassan, H. Naeem, B. Safaei, and Q. Zheng, "IMCFN: Image-based malware classification using fine-tuned convolutional neural network architecture," Comput. netw., vol. 171, no. 107138, p. 107138, 2020.

[19] "Advice for the public on COVID-19 – World Health Organization," Who.int. [Online]. Available: https://www.who.int/emergencies/diseases/novel-coronavirus-2019/advice-for-public

[20] P. Viola and M. Jones, "Rapid object detection using a boosted cascade of simple features," Proceedings of the 2001 IEEE Computer Society Conference on Computer Vision and Pattern Recognition. CVPR 2001, pp. I-I, 2001.

[21] C. Szegedy, W. Liu, Y. Jia, P. Sermanet, S. Reed, D. Anguelov, D. Erhan, V. Vanhoucke, and A. Rabinovich., "Going deeper with convolutions. In Proceedings of the IEEE Conference on Computer Vision and Pattern Recognition", pages 1–9, 2015

[22] C. Szegedy, V. Vanhoucke, S. Ioffe, J. Shlens and Z. Wojna, "Rethinking the Inception Architecture for Computer Vision," 2016 IEEE Conference on Computer Vision and Pattern Recognition (CVPR), pp. 2818-2826, 2016

[23] K. Weiss, T. M. Khoshgoftaar, and D. Wang, "A survey of transfer learning," J. Big Data, vol. 3, no. 1, 2016.

[24] Keras Team, "Keras applications," Keras.io. [Online]. Available: https://keras.io/api/applications/.

[25] X. Liu, C. Wang, Y. Hu, Z. Zeng, J. Bai, and G. Liao, "Transfer learning with convolutional neural network for early gastric cancer classification on magnifiying narrow-band imaging images," In 2018 25th IEEE International Conference on Image Processing (ICIP), pp. 1388-1392. IEEE, 2018.

[26] A. Arjmand, S. Meshgini and R. Afrouzian, Breast tumor detection using convolutional neural network in MRI images, Journal of Advances Signal Processing, vol. 3, no. 2, pp. 109-117, 2019

M. Peyrohoseini nejad, A. Karami, Automatic small defect detection in unmanned aerial vehicle images of power transmission lines using DRSPTL, Journal of Advances Signal Processing, vol. 4, no. 2, pp. 159-170, 2020.

زیرنویس‌ها

[1] Support Vector Machine
[2] Hyper Parameter
[3] Trade off
[4] Fully Connected layers
[5] Validation
[6] AvgPool
[7] Flatten
[8] Dropout
[9] Softmax layer
[10] epoch
[11] Fit
[12] Categorical Cross Entropy
[13] Data augmentation
[14] overfitting
[15] Rotate
[16] Zoom
[17] Shift
[18] Transfer learning
[19] Freeze
[20] Fine-tuning
[21] Recall
[22] Confusion matrix